\documentclass[conference]{IEEEtran}
\usepackage{times}
\usepackage{wrapfig,lipsum,booktabs}
\usepackage[numbers]{natbib}
\usepackage{amsfonts}
\usepackage{algorithm}
\usepackage{algpseudocode}
\usepackage{multicol}
\usepackage{graphicx}
\usepackage[bookmarks=true]{hyperref}
\usepackage{lipsum}
\usepackage{float}
\usepackage{amsmath}
\DeclareMathOperator*{\argmin}{argmin} 
\usepackage{graphicx}
\usepackage{color}
\usepackage{xcolor}

\newcommand\freefootnote[1]{%
  \let\thefootnote\relax%
  \footnotetext{#1}%
  \let\thefootnote\svthefootnote%
}

\begin{document}

\title{Render and Diffuse: Aligning Image and Action Spaces for Diffusion-based Behaviour Cloning}

\author{
\authorblockN{Vitalis Vosylius$^{1, 2, *}$ , Younggyo Seo$^{1}$, Jafar Uruç$^{1}$, Stephen James$^{1}$}
\authorblockA{$^{1}$ Dyson Robot Learning Lab \quad $^{2}$ Imperial College London \\ \href{https://vv19.github.io/render-and-diffuse}{vv19.github.io/render-and-diffuse}}}

\maketitle

\begin{abstract}
In the field of Robot Learning, the complex mapping between high-dimensional observations such as RGB images and low-level robotic actions, two inherently very different spaces, constitutes a complex learning problem, especially with limited amounts of data. In this work, we introduce \textit{Render and Diffuse (R\&D)} a method that unifies low-level robot actions and RGB observations within the image space using virtual renders of the 3D model of the robot. Using this joint observation-action representation it computes low-level robot actions using a learnt diffusion process that iteratively updates the virtual renders of the robot. This space unification simplifies the learning problem and introduces inductive biases that are crucial for sample efficiency and spatial generalisation. We thoroughly evaluate several variants of \textit{R\&D} in simulation and showcase their applicability on six everyday tasks in the real world. Our results show that \textit{R\&D} exhibits strong spatial generalisation capabilities and is more sample efficient than more common image-to-action methods.

\end{abstract}

\IEEEpeerreviewmaketitle

\section{Introduction}
\label{sec:intro}

\freefootnote{$^{*}$Work done during an internship at the Dyson Robot Learning Lab.}

When learning to predict actions from high-dimensional inputs (e.g. RGB images), a neural network must learn a highly complex mapping between two inherently different spaces. 
Recent research~\cite{james2022coarse, shridhar2023perceiver} has shown that predicting actions in the same space as observations leads to a drastic increase in sample efficiency and spatial generalisation capabilities. 
However, such methods typically utilise 3D representations that rely on accurate depth information and operate hierarchically by predicting the next-best pose that is reached open-loop either using a motion planner~\cite{james2022coarse, shridhar2023perceiver, gervet2023act3d} or a separate learned policy~\cite{ma2024hierarchical, xian2023chaineddiffuser}.

\begin{figure}[t!]
  \centering
  \includegraphics[width=\columnwidth]{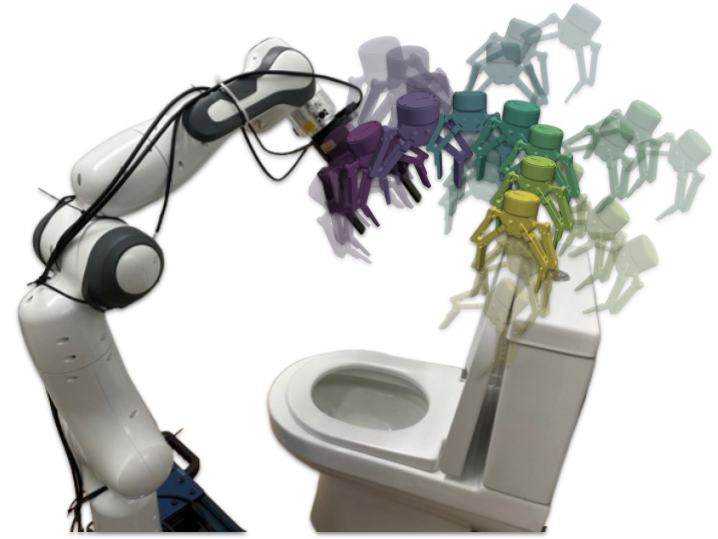} 
  \caption{A high-level overview of \textit{Render and Diffuse} process in the context of a putting the toilet seat down task. Poses of the renders of the robot gripper are updated iteratively using a learnt denoising process.}
  \label{fig:fig_1}
\end{figure}

In this work, we aim to leverage insights from the aforementioned methods, but instead of relying on 3D representations and next-best pose action formulation, we learn low-level control policies from RGB images while still aligning observation and action spaces.
Representing low-level actions within the observation space allows us to simplify the learning problem and increase the sample efficiency and spatial generalisation capabilities of the learnt policies. We do so using a process we call \textit{Render and Diffuse (R\&D)} and represent low-level robotic actions within the image space by rendering the robot in the configuration that would be achieved if the considered actions were to be taken. By utilising a learnt action denoising process (similar to \cite{chi2023diffusion_policy}), such action representations can be updated by iteratively rendering the robot in different configurations until they represent actions that closely align with those in the training data (see Figure~\ref{fig:fig_1}). Several different options of mapping our devised rendered action representation back to low-level robot actions result in a family of \textit{R\&D} methods that unify image and action spaces in unique ways. 

This observation-action unification through virtual renders of the robot allows the model to understand the spatial implications of the considered actions and introduces inductive biases that are crucial for sample-efficient spatial generalisation. Our contributions are three-fold: 
 
\textbf{1)} We propose a novel way of combining low-level actions and RGB observations within a unified image space.
 
\textbf{2)} We present a family of \textit{R\&D} methods that utilise this representation and introduce different ways of iteratively updating low-level actions using a denoising process learnt from demonstrations.
 
\textbf{3)} We thoroughly evaluate our proposed method in simulation, systematically studying its spatial generalisation capabilities and sample efficiency, as well as showcasing its capabilities on a variety of real-world tasks.

\section{Related Work}
\label{sec:rel_work}
\textbf{Aligning Observation and Action Spaces.} Robotic control policy learning is typically approached by mapping observations to actions \cite{zhao2023aloha, jang2022bc, chi2023diffusion_policy}. Due to the complexity of this mapping, learning it without introducing biases is challenging. Recently, methods that represent actions in the same space as the observations have shown tremendous success, especially in the context of spatial generalisation and sample efficiency. For instance, methods like C2F-ARM~\cite{james2022coarse} and PerAct~\cite{shridhar2023perceiver} voxelise point cloud observations for next-best-pose predictions within the same voxelised space. 
Act3D~\cite{gervet2023act3d} and ChainedDiffuser~\cite{xian2023chaineddiffuser} use point cloud observations and candidate points sampled in the same Cartesian space to make next-best-pose predictions.
Similarly, Implicit Graph Alignment~\cite{vosylius2023few} unifies observations and actions in a graph representation derived from point cloud data \cite{vosylius2023few, vosylius2023start}. Transporter Nets~\cite{zeng2021transporter} and CLIPort~\cite{shridhar2022cliport} employ feature maps learned from RGBD observations and template matching to identify pick-and-place poses. Additionally, many methods have utilised optical~\cite{argus2020flowcontrol, weng2022fabricflownet} or 3D~\cite{seita2023toolflownet, fu2023pt, zhou2023learning} flow predictions in learning to complete a variety of tasks. However, all these methods are either restricted to simple tabletop environments, require explicit access to depth information, or are not suited for learning low-level robotic action as they utilise discretised next-best-pose predictions. In contrast, our approach seeks to align RGB-only observations with low-level robotic actions. We achieve this by using renderings of the robot in configurations that would be reached through the execution of these low-level actions. This technique offers a novel way to integrate observations and actions without the constraints observed in previous methods.

\textbf{Diffusion Models in Robotics.} Diffusion Models have recently gained traction in robotic applications, being employed in diverse and innovative ways. For instance, image Diffusion Models \cite{mandi2022cacti, mandlekar2023mimicgen, yu2023scaling} have been used to create image augmentations, helping robots to adapt to varying environments with distractor objects and different visual settings. Image Diffusion Models have also been used to `imagine' object rearrangement goals~\cite{kapelyukh2023dall} or subgoals~\cite{black2023zero} for low-level robotic policies. Diffusion Policy~\cite{chi2023diffusion_policy} applies diffusion models to visuomotor policy learning, demonstrating strong results in solving complex real-world robotic tasks. Subsequently, many works have utilised similar formulations of learning the conditional distribution of actions for low-level control \cite{chen2023playfusion, reuss2023goal, ha2023scaling} using Diffusion Models. Contrary to these approaches, we propose the use of Diffusion Models within a unified Observation-Action space. This strategy eliminates the complexity of learning intricate mappings between separate observation and action spaces, simplifying the learning process and enhancing its sample efficiency and generalisation capabilities.

\section{Preliminaries} 
\label{sec:preliminaries}

\textbf{Diffusion Models.} Diffusion models are a class of generative models that employ a forward and backward Markov-chain diffusion process to effectively capture complex and multimodal data distributions. In the forward phase, these models iteratively add noise to a sample \( x^0 \) extracted from the real data distribution \( q(x) \), as described by~\cite{ho2020ddpm}:
\begin{equation}
    \label{eq:forward_diffusion}
    q(x^k \mid x^{k-1}) = \mathcal{N}(x^k; \sqrt{1-\beta_k}x^{k-1}, \beta_k \mathbf{I}), \quad k = 1, \ldots, K
\end{equation}
Here, \( \mathcal{N} \) represents the normal distribution, \( \beta_k \) the variance schedule, and \( K \) the total number of diffusion steps. This process gradually transitions the data sample into a Gaussian noise distribution as \( K \) increases.

Inversely, in the reverse diffusion process, the aim is to reconstruct the original data from its noise-altered state, utilising a parameterised model \( p_{\theta}(x^{k-1} \mid x^k) \). This model is trained to estimate the reverse conditional distribution of the forward process, thereby learning to generate new samples starting from a normal distribution \( \mathcal{N} \) via an iterative process:

\begin{equation}
    p_{\theta}(x^{k-1} \mid x^k) = \mathcal{N}(x^{k-1}; \mu_{\theta}(x^k, k), \Sigma_{\theta}(x^k, k))
\end{equation}

Here, \( \mu_{\theta}(x^k, k) \) and \( \Sigma_{\theta}(x^k, k) \) are functions parameterised by \( \theta \), predicting the mean and covariance at each diffusion step \( k \). The training process, which focuses on optimising the variational lower bound of the data likelihood, has been shown to generate high-quality samples in various applications, including image and audio generation \cite{rombach2022high, ramesh2021zero, schneider2023archisound}.

Recently, diffusion models have also been successfully applied to control problems by learning a conditional distribution of control parameters $\mathbf{a}$ as $p_{\theta}(\mathbf{a}^{k-1} | \mathbf{a}^{k}, o)$, where $o$ is an observation of the environment. In this work, we build on top of such a formulation and extend it by introducing a novel way to fuse $\mathbf{a}^{k}$ and $o$ as well as update $\mathbf{a}^{k}$ based on the predictions made in the observation space.

\section{Render and Diffuse} 
\label{sec:method}

\textbf{Problem Setting}. We consider a standard behaviour cloning setting, where given a dataset of demonstrations as RGB image and action pairs $D = {\left(I, \mathbf{a}\right)_{i=1}^N}$, with $I \in \mathcal{I}$ and $\mathbf{a} \in \mathcal{A}$, we aim to obtain a policy that maps images to actions: $\pi: \mathcal{I} \to \mathcal{A}$. The primary difficulty in deriving a policy that maps images to actions is the intrinsic disparity between their representational spaces. Learning this mapping with limited data while ensuring spatial generalisation presents a significant challenge.

\textbf{Overview.} We introduce a strategy to unify image and action domains within a single image space $\mathcal{I}$, simplifying the learning process by reducing the complexity inherent in navigating between these fundamentally different spaces. We utilise a rendering procedure and a known model of the robot to visually `imagine' the potential spatial implications of the robot's considered actions. Through a learned denoising process, these rendered actions are updated until they closely align with the actions observed in our dataset $D$. A high-level overview of our approach can be seen in Figure~\ref{fig:fig_1}.

\subsection{Rendered Action Representation}
\label{sec:representation}

\begin{figure}[ht!]
  \centering
  \includegraphics[width=\columnwidth]{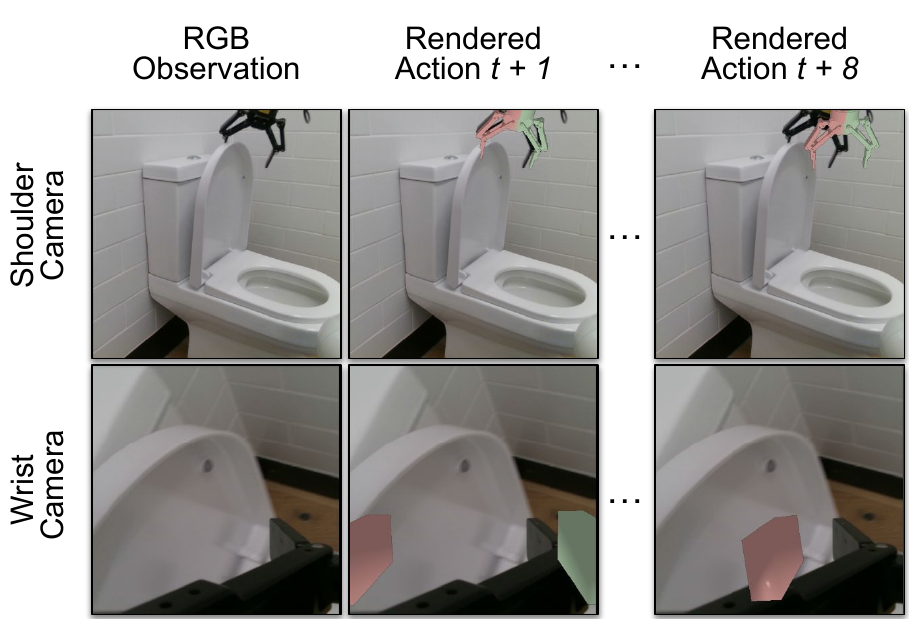} 
  \caption{Visualisation of the RGB observations from two different cameras and rendered action representation at time steps $t+1$ and $t+8$ for a task of putting the toilet seat down.}
  \label{fig:ra}
\end{figure}

The initial phase of our approach involves integrating state observations with actions in a unified image space. Generally, predicting state changes in response to specific actions requires an accurate model of the environment, including its dynamics. Acquiring or learning such a model is often impractical or infeasible. However, 3D models of robots and their kinematics, along with camera models, are widely available. Utilising these resources, we propose to represent robot actions as rendered images of the robot in a configuration that would be kinematically achieved if the intended action was executed. In this way, we represent observations (RGB images) and low-level robot actions (e.g. end-effector velocities) in the same image space by `imagining' the spatial implications of robot actions on its embodiment, while leaving the learning of the environment's dynamics to the downstream model.
In this work, we consider actions as end-effector velocities and render only an open gripper, although this approach could be readily expanded to encompass the robot's full configuration using Forward Kinematics. We then overlay this rendering onto the current RGB image, as illustrated in Figure~\ref{fig:ra}. Note that for wrist-mounted cameras, we render only the gripper's fingers to prevent visual obstructions. 
This is done using a different 3D model, representing just the fingers of the gripper.
Additionally, to break the symmetry that is present in most two-fingered grippers, we use a colour texture to aid the downstream model in easily distinguishing its 6D pose in the image.

\textbf{Rendering Procedure.} To determine the pose in which the gripper should be rendered, we first calculate its potential spatial pose in the frame of the robot's base following an action $\mathbf{a}$, expressed as $T_{w\_a} = T_{w\_g} \times T(\mathbf{a})$. Here, $T(\mathbf{a}) \in \mathbb{SE}(3)$ denotes a relative transformation representing the intended action. $T(\mathbf{a})$ is constructed from end-effector velocities ($\mathbf{a} \in se(3)$) through an exponential mapping: $T(\mathbf{a}) = \text{Expmap}(\mathbf{a})$. Then, by utilising the camera $c$'s extrinsic matrix $T_{w\_c}$, we can reposition the gripper's CAD model in the camera's frame and render an image of it, creating the rendered action representation, which we denote as $R^c$:

\begin{equation}
    R^c, P^c = Render(T^{-1}_{w\_c} \times T_{w\_g} \times T(\mathbf{a}) \times Gripper, K^c)
    \label{eq:rendered_actions}
\end{equation}

Here, $K^c$ is the intrinsic matrix of the camera. In addition, to $R^c$ we also have access to the partial point cloud of the rendered gripper $P^c$ as the rendering procedure provides us with ground truth depth information. Such a procedure can be done for an arbitrary number of cameras $c$ as long as their extrinsic and intrinsic matrices are known. For brevity, we omit the superscript $c$ in subsequent discussions except when explicitly talking about multiple cameras.

\begin{figure}[ht!]
  \centering
  \includegraphics[width=\columnwidth]{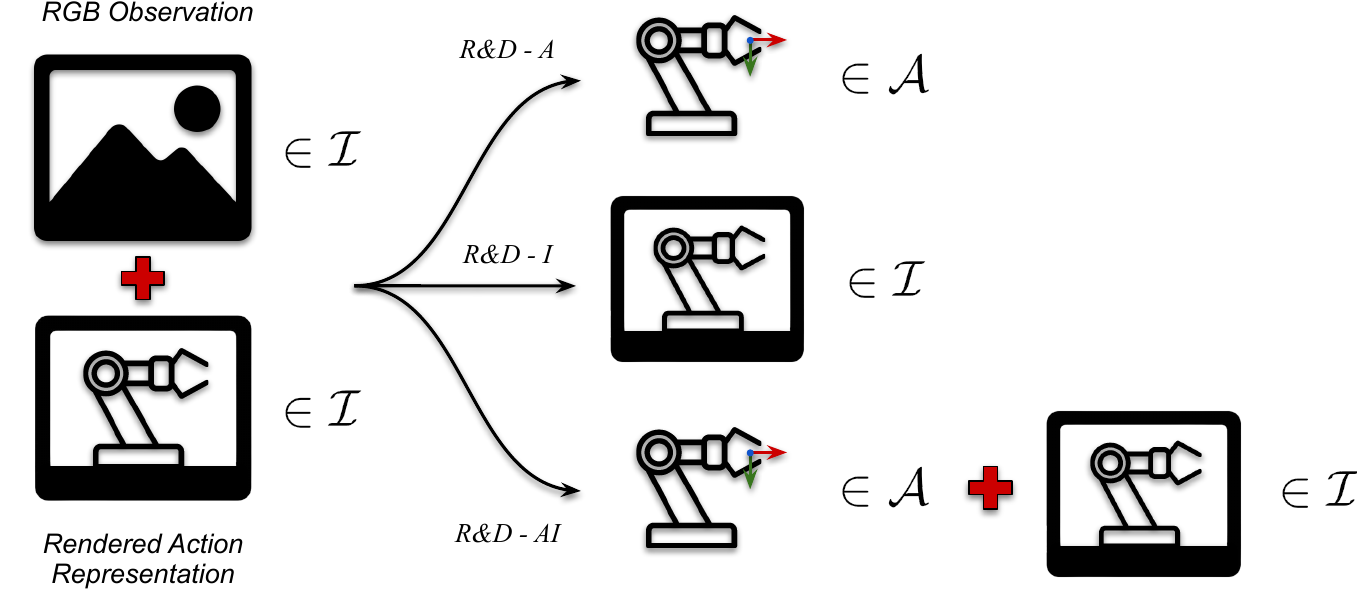} 
  \caption{An overview of the members of \textit{R\&D} method family, each combining observation ($\mathcal{I}$) and action ($\mathcal{A}$) spaces in a unique way.}
  \label{fig:variants}
\end{figure}

\begin{figure*}[ht!]
  \centering
  \includegraphics[width=\textwidth]{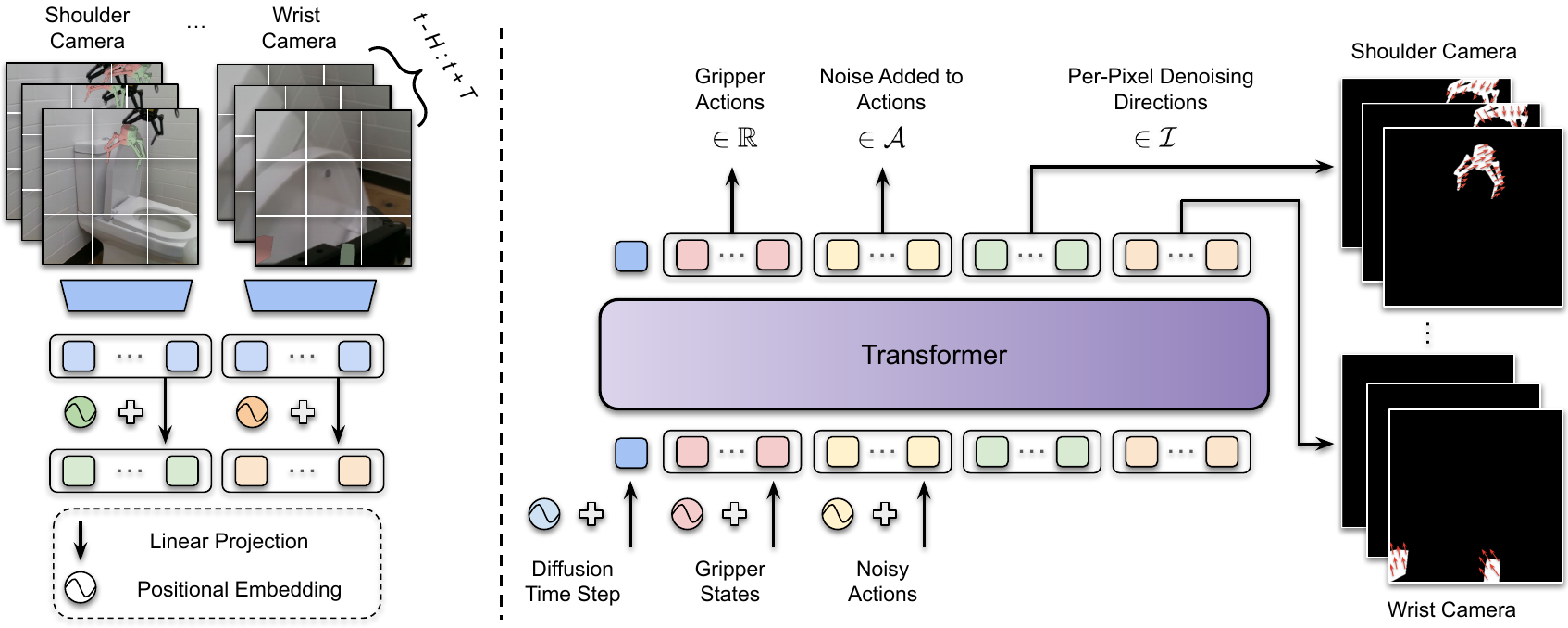}
  \caption{Overview of the architecture of the transformer used in the \textit{R\&D} process. Different positional embeddings are added to the tokenized input that is fed into a Transformer model. Token embeddings processed with several self-attention layers are then decoded into per-camera denoising directions (image space), binary gripper actions and noise added to the ground truth actions in the action space.}
  \label{fig:architecture}
\end{figure*}

Even if we can express actions and observations in a unified space as $R$, we still need a method to systematically update them until they closely align with the actions observed in our dataset $D$. To this end, we utilise the diffusion models described in Section~\ref{sec:preliminaries} and learn how the rendered representation of noisy actions should be altered such that it becomes closer to the training data distribution. We call this process \textit{Render and Diffuse (R\&D)}.

\subsection{R\&D Family:}
\label{sec:variants}

With our rendered action representation $R$ as an input to the model, we must also decide how we are going to update it, i.e., what our learned model should predict and in which space (image or action). 
This decision gives rise to several \textit{R\&D} variants forming a family of methods (see Figure~\ref{fig:variants}), each of which we describe below.
In this work, we treat the gripper's opening-closing actions as a separate binary variable $a_g$, and predict it independently for all variants of \textit{R\&D} (\textit{Gripper Actions} in Figure~\ref{fig:architecture}). We do so because we render only an open gripper as we mentioned in Section~\ref{sec:representation}.

\textbf{R\&D-A}. The most straightforward way to learn how to update $R$ is by learning the noise that was added to the ground truth actions directly in the action space $\mathcal{A}$, as done in~\cite{chi2023diffusion_policy}. In such a case, the function that we would effectively learn is $p_\theta(\textbf{a}^{k-1} \mid R(\textbf{a}^{k}))$. The model would have explicit access to the spatial implications of the currently considered actions $\textbf{a}^{k}$ through $R(\textbf{a}^{k}))$ and would be tasked with determining how to update them. We hypothesise that this approach can greatly aid the model in learning the mapping between observation and action spaces using this newly introduced inductive bias.

\textbf{R\&D-I}. Another option, which would further align observation and action spaces is to predict the denoising direction in the same image space $\mathcal{I}$. We represent this denoising direction as $F \in \mathbb{R}^{H \times W \times 3}$, where each pixel value of $F$ holds information on in which direction a specific part of the robot should move to get closer to the actions represented in the dataset $D$. It can be viewed as a spatial gradient or a per-pixel 3D flow expressed in the camera frame. 
Thus, in this case, the function we aim to learn (including the gripper actions $a_g$) can be expressed as:

\begin{equation}
    F, a_g = f_\theta \left(\{I, R(\mathbf{a_{noisy}}), s_g, k\right).
    \label{eq:flow_learning}
\end{equation}

Here, $I$ is the RGB observation, $R(\mathbf{a_{noisy}})$ is the rendered representation of the noisy action, $s_g$ is the current state of the gripper, and $k$ is the diffusion time step.

\textbf{R\&D-AI}. In theory, completely aligning action and observation spaces by predicting the denoising direction as $F$ should simplify the learning problem the most, however, several considerations must be made here. First, renders of the gripper may happen to be outside of all the camera views, and second, learning the action distribution with high precision in such a high-dimensional output space can be challenging. To address these, we propose to combine the previously described $R\&D-A$ and $R\&D-I$ approaches and make predictions in both action $\mathcal{A}$ and image $\mathcal{I}$ spaces. In cases where $R$ is outside the image, we can then rely on predictions made directly in the action space. Additionally, we can use predictions in the action space at the last diffusion step, a part of the denoising process responsible for high-frequency, precise refinements, to increase the precision of our predicted actions. 

\subsection{Learning Problem}
\label{sec:learning}
\textbf{History and Prediction Horizons.} In line with~\cite{chi2023diffusion_policy}, we aim to predict $T$ future actions and include $H$ past observations in the learning problem. Additionally, we include images from multiple cameras ($C$) and make predictions $F$ for each camera individually. Thus, we are learning to predict $\{F_c^{t:t+T}\}^{c \in C}, \mathbf{\epsilon}^{t:t+T}, a_g^{t:t+T}$, where $t$ represents the current time step in a trajectory, and $\mathbf{\epsilon}$ is the noise that was added to the actions as described in Section~\ref{sec:preliminaries}. For clarity, we henceforth omitted all the superscripts in our subsequent discussions. 

\textbf{Training.} We learn $f_\theta$ by minimising the following:

\begin{equation}
    Loss = L1(F, F^{gt}) + L1(\mathbf{\epsilon}, \mathbf{\epsilon}^{gt}) + BCE(a_g, a^{gt}_g)
    \label{eq:Loss}
\end{equation}

Here, $F^{gt}$ and $a^{gt}_g$ are ground truth denoising directions and gripper actions. The first two terms in Equation~\ref{eq:Loss} are responsible for learning the diffusion denoising direction in image and action spaces, respectively, while the last term accounts for binary gripper action predictions. We utilise $L1$ loss instead of $MSE$ as it has been shown to reduce deviations from the data distribution in some applications~\cite{saharia2022palette}. We obtain ground truth denoising direction as: 

\begin{equation}
    F_{gt} = T(\mathbf{a_{gt}}, \mathbf{a_{noisy}}) \times P - P.
    \label{eq:gt_flow}
\end{equation}

Here, $T(\mathbf{a_{gt}}, \mathbf{a_{noisy}}) \in \mathbb{SE}(3)$ is a transformation that adjusts the gripper from its pose resulting from $\mathbf{a_{noisy}}$ to the pose associated with the ground truth action in the camera frame. $F_{gt}$, therefore, represents the difference between where each point of the rendered gripper should be according to ground truth actions and where they ended up due to noisy actions. Note that if a full robot configuration is used, a set of transformations $T(\mathbf{a_{gt}}, \mathbf{a_{noisy}})$ can be obtained using Forward Kinematics. A complete training procedure is described in Algorithm~\ref{alg:training}.

\begin{algorithm}
\caption{Training Procedure}\label{alg:training}
\begin{algorithmic}[1] 

\For{$i = 1$ \textbf{to} \textit{TrainingSteps}}
    \State $k \gets \text{Randint}(0, K)$ \Comment{Sample diffusion time step}
    \State $(I, s_g, \mathbf{a}_{gt}, g_{gt}) \sim D$ \Comment{Sample the dataset}
    \State $\epsilon_{gt} \gets \mathcal{N}(0, \mathbf{I})$ \Comment{Sample noise}
    \State $\mathbf{a}_{noisy} \gets \text{AddNoise}(\mathbf{a}_{gt}, \epsilon_{gt}, k)$ \Comment{Eq.~\ref{eq:forward_diffusion}}
    \State $(R, P) \gets \text{Render}(\mathbf{a}_{noisy})$ 
    \State $(F, \epsilon, g) \gets f_\theta(I, s_g, RA, k)$ 
    \State $F_{gt} \gets \text{ComputeLabels}(P, \mathbf{a}_{noisy}, \mathbf{a}_{gt})$ \Comment{Eq.~\ref{eq:gt_flow}}
    \State $\text{Loss} \gets \text{L1}(F_{gt}, F) + \text{L1}(\epsilon_{gt}, \epsilon)  + \text{BCE}(g_{gt}, g)$
    \State $\theta \gets \theta - \alpha \nabla_{\theta} \text{Loss}$ \Comment{Update network weights}
\EndFor

\end{algorithmic}
\end{algorithm}

\subsection{Architecture}
\label{sec:Architecture}

We employ a Vision Transformer (ViT)~\cite{dosovitskiy2020image} style architecture, as illustrated in Figure~\ref{fig:architecture}, to learn the denoising function $f_\theta$. The Transformer architecture is chosen for its efficiency in fusing information from various sources. This includes data from different time steps (both past and future), multiple camera viewpoints, and other conditional variables such as the states of the gripper and the diffusion time step. 

\textbf{Network Structure Overview}. First, patches of image observations and the rendered action representations are converted into tokens (Figure~\ref{fig:architecture} left). Embeddings of gripper state information, diffusion time step and action tokens are then concatenated to the image tokens and processed with several self-attention layers (middle). To enable the network to distinguish information coming from multiple camera views and different time steps, different learnable positional embeddings are added to the corresponding tokens. Finally, corresponding token embeddings are decoded (via linear projection) into per-camera view denoising direction predictions $F$ (right), gripper actions $a_g$, and noise $\epsilon$ that was added to the ground truth actions expressed in the action space $\mathcal{A}$. 
Note that R\&D-AI makes predictions in both image and action spaces, while R\&D-A and R\&D-I variants do not predict per-pixel denoising directions ($F \in \mathcal{I}$) and noise added to the actions ($\epsilon \in \mathcal{A}$), respectively.

\subsection{Inference}
\label{sec:inference}

\begin{figure}[ht!]
  \centering
  \includegraphics[width=\columnwidth]{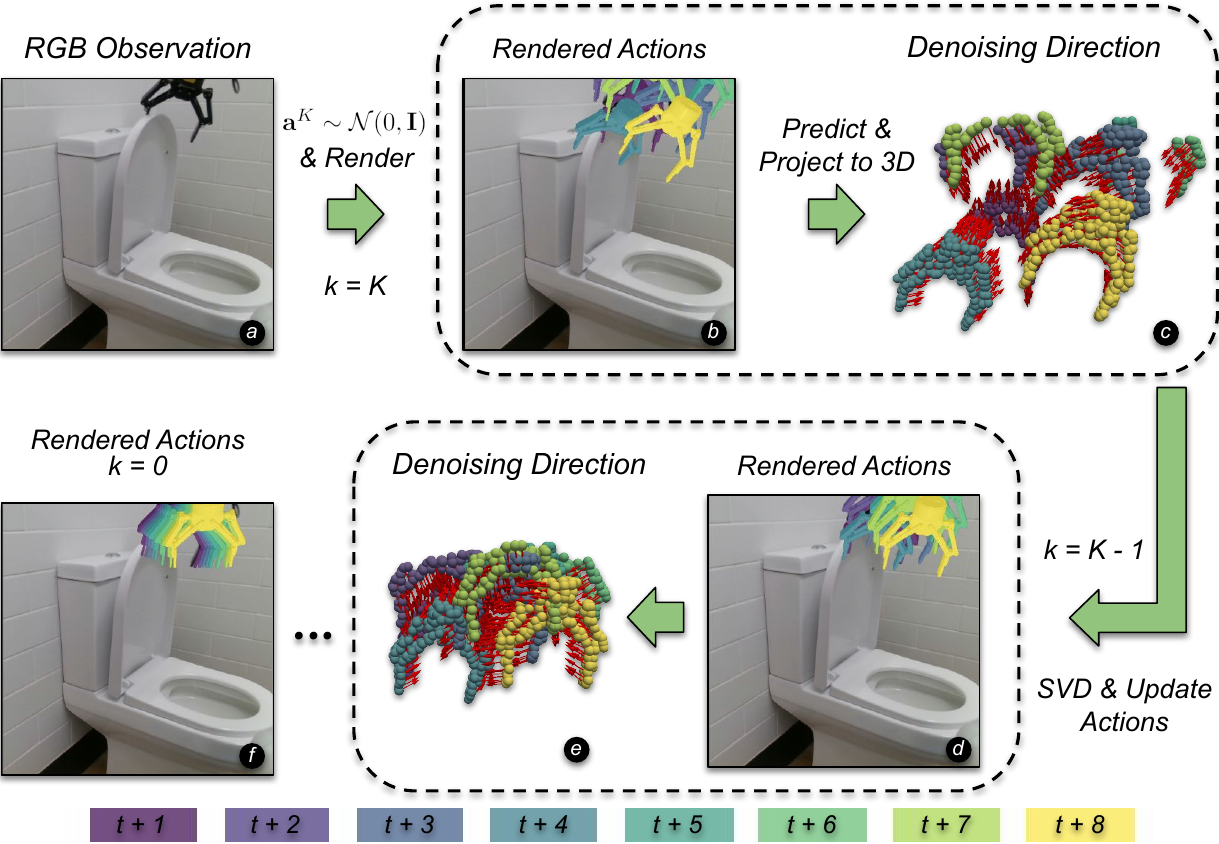}
  \caption{Overview of the \textit{Render and Diffuse} inference process showing (a) an initial RGB observation and (b, e, f) rendered action representations and (c, d) partial point cloud of the renders at different diffusion time steps. Different colours indicate actions at different time steps in the future.}
  \label{fig:inf}
\end{figure}

Figure~\ref{fig:inf} shows the overview of the \textit{Render \& Diffuse} process at inference. It starts with an RGB observation of the environment (a), initialising actions $\mathbf{a}_k$ from a normal distribution $\mathcal{N}(0, \mathbf{I})$ and constructing the rendered action representations $R^k$, together with a partial point cloud $P^k$ of the rendered grippers (b). This is then used to make per-camera view denoising direction  $\{F_c^k\}^{c \in C}$ and gripper actions $a_g$ predictions. Knowing the direction in which each point on the rendered gripper should move ($F^k$), we then update the positions of the gripper’s point cloud ($P^k$) by taking a denoising step according to the DDIM~\cite{song2020ddim} (c):

\begin{equation}
    P^{k-1} = \sqrt{\alpha_{k-1}} \hat{P}^0 + \sqrt{\frac{1 - \alpha_{k-1}}{1 - \alpha_{k}}} \left(P^{k} - \sqrt{\alpha_k} \hat{P}^0 \right)
\end{equation}

Here, $\hat{P}^0=P^k + F^k$ is the estimated positions of the points at $k=0$.
We then aggregate point clouds coming from different camera views by projecting them into a common frame of reference. This leaves us with 2 point clouds $P^{k-1}$ and $P^{k}$, that implicitly represent gripper poses at denoising time step $k$ and $k-1$. As we know the ground truth correspondences between them, we can extract an $\mathbb{SE}(3)$ transformation that would align them using a Singular Value Decomposition (SVD) as~\cite{arun1987least}:

\begin{equation}
    T_{k-1, k} = \argmin_{T_{k, k-1} \in \mathbb{SE}(3)} ||P^{k - 1} - T_{k-1, k} \times P^{k}||^2
    \label{eq:registration}
\end{equation}

In such a way, predictions from multiple views are fused without requiring the network to learn the camera extrinsics implicitly. Finally, the $\mathbf{a^{k-1}}$ is calculated by applying calculated transformation $T_{k-1, k}$ to $\mathbf{a^{k}}$. This process is repeated $K$ times (d-e) until the final $\mathbf{a^{0}}$ actions are extracted (f).

\section{Experiments} 
\label{sec:experiments}

To thoroughly assess the effectiveness of \textit{R\&D} in learning low-level control policies from a limited number of demonstrations, and its spatial generalisation capabilities and practicality in real-world robotic tasks, we conducted experiments in three distinct settings: (1) a simulation setting, where we compare \textit{R\&D} against state-of-the-art behaviour cloning methods on $11$ different tasks from RLBench~\cite{james2020rlbench}, (2) a systematic evaluation of interpolation inside the convex hull of the demonstration on four RLBench tasks, where significant spatial generalisation is required, and (3) a real-world setting, where we evaluate the performance of the complete robotic pipeline on six everyday tasks. Additionally, we show \textit{R\&D} working in a multi-task setting, and perform an ablation study on the hyper-parameters of the transformer model to further justify our design choices. 

\subsection{Baselines}
\label{sec:baselines}

We compare \textbf{R\&D} against state-of-the-art behaviour cloning methods that have recently shown impressive results in learning control policies for complex robotic tasks. Namely, use \textbf{ACT}~\cite{zhao2023aloha} and \textbf{Diffusion Policy (DP)}~\cite{chi2023diffusion_policy} as baselines. 
Both ACT and Difucion Policy, use the ResNet-18 vision backbone as proposed in the original implementations. Similar to \textit{R\&D}, ACT utilises a transformer architecture while Diffusion Policy uses a convolutional neural network. Hyper-parameters such as prediction horizon, observation history or the number of trainable parameters were adjusted to match that of \textit{R\&D} for fair comparison.
We would like to note that by comparing against such baselines that have already shown strong performance on complex tasks, our goal is not to beat them in terms of performance in large-data regimes or task complexity. Instead, we aim to use them to showcase that representing low-level robot action inside the image space can improve spatial generalisation and performance in low-data regimes.
In addition to the aforementioned baselines, we test three variants of our method \textbf{R\&D-A}, \textbf{R\&D-I}, and \textbf{R\&D-AI}, described in Section~\ref{sec:variants}.

\subsection{Hyper-parameters}
\label{sec:params}

\textbf{Common parameters.} For a fair comparison, all methods, including baselines, use RGB observations ($128 \times 128$) from one external camera and one camera mounted on the wrist of the robot. Additionally, all methods use $H=2$ last observations as input and predict $T=8$ future actions. Actions are represented as relative end-effector displacements, an actions space which we found to work the best for all the methods in the simulation setting. All models have been trained for $100k$ iterations (with a batch size of $8$) and roughly have the same number of trainable parameters ($90M - 100M$).

\textbf{R\&D parameters.} We use a Transformer consistent of $8$ self-attention layers, embedding hidden dimension of $1024$, $16$ attention heads, and a patch size of $16$. During training, we use $50$ diffusion denoising steps while at inference we only take $4$, which is possible due to the use of DDIM noise scheduler~\cite{song2020ddim}. We are able to run the whole denoising process at $4$HZ on a laptop with an NVIDIA RTX A3000 GPU without extensive optimisation.

\subsection{Evaluation in Simulated Environments}
\label{sec:sim_eval}

Our first set of simulated experiments aims to investigate our model's capability of completing various robotic manipulation tasks with limited amounts of demonstrations.

\textbf{Experimental Procedure.} We use a standard RLBench demonstration collection procedure using the Franka Emika Panda robot and collect $20$ demonstrations for $11$ different tasks. For this set of experiments, objects in the environment are initialised randomly, a strategy which does not guarantee a good coverage of the workspace with a small number of demonstrations.
Additionally, demonstrations are a combination of the trajectories of linear end-effector movements and paths produced by RRT-Connect~\cite{kuffner2000rrt} planner resulting in a multimodal dataset of demonstrations. For the \textit{Place Phone on Base} and \textit{Slide Block to Target} tasks, we had to restrict the rotation of the objects between $-45$ and $45$ degrees, otherwise, paths produced by the RRT planner would have dominated the dataset removing any correlation between the demonstration. 
We also set the starting configuration of the robot, such that the whole gripper is in view from the front camera. We did this to be able to study the performance of one of \textit{R\&D} variants that relies solely on the predictions made in the image space - \textit{R\&D-I}.

We train our model and the baselines on $20$ collected demonstrations as well as on a smaller subset of $10$ demonstrations. We evaluate trained models by initialising objects in the environment in a $100$ different poses that have not been seen during the demonstration collection.

\begin{table*}[]
\resizebox{\textwidth}{!}{%
\begin{tabular}{l|cc|cc|cc|cc|cc|cc|cc|cc|cc|cc|cc|cc}
 & \multicolumn{2}{c|}{\begin{tabular}[c]{@{}c@{}}Reach \\ Target\end{tabular}} & \multicolumn{2}{c|}{\begin{tabular}[c]{@{}c@{}}Push \\ Button\end{tabular}} & \multicolumn{2}{c|}{\begin{tabular}[c]{@{}c@{}}Close \\ Microwave\end{tabular}} & \multicolumn{2}{c|}{\begin{tabular}[c]{@{}c@{}}Pick \\ up Cup\end{tabular}} & \multicolumn{2}{c|}{\begin{tabular}[c]{@{}c@{}}Push \\ Buttons\end{tabular}} & \multicolumn{2}{c|}{\begin{tabular}[c]{@{}c@{}}Close \\ Laptop\end{tabular}} & \multicolumn{2}{c|}{\begin{tabular}[c]{@{}c@{}}Phone\\ on Base\end{tabular}} & \multicolumn{2}{c|}{\begin{tabular}[c]{@{}c@{}}Slide\\ Block\end{tabular}} & \multicolumn{2}{c|}{\begin{tabular}[c]{@{}c@{}}Lift \\ Lid\end{tabular}} & \multicolumn{2}{c|}{\begin{tabular}[c]{@{}c@{}}Open \\ Box\end{tabular}} & \multicolumn{2}{c|}{\begin{tabular}[c]{@{}c@{}}Open \\ Drawer\end{tabular}} & \multicolumn{2}{c}{Average} \\[0.3cm] 
\# of demos & \#10 & \#20 & \#10 & \#20 & \#10 & \#20 & \#10 & \#20 & \#10 & \#20 & \#10 & \#20 & \#10 & \#20 & \#10 & \#20 & \#10 & \#20 & \#10 & \#20 & \#10 & \#20 & \#10 & \#20 \\ \toprule
DP & 8 & 29 & 39 & 70 & 62 & 87 & 29 & 46 & 4 & 31 & 35 & 70 & 18 & 41 & 32 & 40 & 63 & 81 & 21 & 39 & \textbf{42} & 45 & 32.1 & 52.6 \\
ACT & 11 & 36 & 43 & 71 & 59 & 81 & 34 & 50 & 9 & 23 & 31 & 64 & 32 & 50 & 41 & \textbf{58} & 70 & 89 & 23 & 46 & 32 & \textbf{50} & 35 & 56.2 \\[0.1cm] \hline 
R\&D-A & 28 & 49 & 52 & 65 & 53 & 87 & 65 & 81 & 69 & 73 & 43 & 73 & 47 & 61 & 35 & 47 & 68 & 89 & \textbf{26} & \textbf{51} & 28 & 27 & 46.7 & 63.9 \\
R\&D-I & 31 & 53 & 55 & 61 & 59 & 88 & 53 & 75 & \textbf{82} & \textbf{88} & 37 & 68 & 50 & 58 & 38 & 51 & \textbf{71} & 90 & 18 & 38 & 15 & 21 & 46.3 & 62.8 \\
R\&D-AI & \textbf{36} & \textbf{61} & \textbf{61} & \textbf{77} & \textbf{65} & \textbf{91} & \textbf{67} & \textbf{84} & 79 & 82 & \textbf{44} & \textbf{71} & \textbf{51} & \textbf{63} & \textbf{43} & 55 & \textbf{71} & \textbf{92} & 25 & 47 & 31 & 32 & \textbf{52.1} & \textbf{68.6}
\end{tabular}
}
\caption{Success rates (\%) of the baselines (Diffusion Policy~\cite{chi2023diffusion_policy} (DP) and Action Chunking with Transformers~\cite{zhao2023aloha} (ACT)) and three variants of \textit{R\&D} evaluated on $11$ RLBench tasks ($100$ test runs each).}
\label{tab:exp1}
\end{table*}

\textbf{Results \& Discussion.} The results for this set of experiments are shown in Table~\ref{tab:exp1}. As evident from the results, \textit{R\&D} outperforms the baselines that do not explicitly combine observation and action spaces. This is particularly noticeable in tasks that require significant spatial generalisation (e.g., \textit{Lift Lid of Saucepan}) or have similarly looking distractors (e.g., \textit{Push Buttons}). This validates our hypothesis that combining observation and action spaces introduces a useful inductive bias, helping the model to better understand the spatial implications of the predicted actions (improving spatial generalisation) and more easily distinguish the task objective (generalising to distractors). On the other hand, methods that learn to map observation to actions directly and lack explicit access to such information need to represent it implicitly, a challenging task in low-data regimes. Furthermore, we observe that \textit{R\&D-I} and \textit{R\&D-A} perform similarly, indicating that most of the performance gain comes from the combination of observations and actions as inputs to the model, rather than the space in which the predictions are made. The \textit{R\&D-AI} variant, which combines predictions from both image and action spaces, performs on average better than the other two variants, justifying the integration of predictions from different sources.

\textbf{Failure Cases of the Baselines} In our experiments, we observed several common failure modes of the evaluated baselines. Firstly, they were prone to overfitting to the proprioception information, ignoring the RGB observations. This resulted in the policies progressing with the task even when the current stage (e.g. grasping an object) is not completed yet. Secondly, imperfect RRT demonstrations from RLBench often negatively impact performance, with policies occasionally unnecessarily committing to intricate movements, taking them out of distribution. Furthermore, both ACT and DP heavily relied on observations from the wrist camera, drifting out of distribution when it could not see the object of interest well. This was particularly critical since we initialised the robot in a configuration that limited the wrist camera's field of view.

\textbf{Failure Cases of \textit{R\&D}.} A clear outlier in Table~\ref{tab:exp1} is the \textit{Open Drawer} task, where \textit{R\&D} performs worse than the baselines. We hypothesise that this is due to several reasons. Firstly, the front camera (the only external camera used) can be heavily occluded by the size and position of the drawer. Secondly, a significant portion of the demonstration trajectories for this task are generated using the RRT planner, resulting in many data samples featuring arbitrary gripper movements relative to the object of interest, i.e., the drawer's handle. As \textit{R\&D} heavily relies on the joint Image-Action representation, this can lead to ambiguities and prevent the model from efficiently learning the task-relevant parts of the image and how the rendered gripper should be moved with respect to them. These issues are less prevalent for the baseline methods, as they rely on global image embeddings and proprioception. To validate our hypothesis, we conducted an experiment where the drawer was always placed in a position such that the front camera could see the drawer's handle and demonstration trajectories did not involve significant movements in arbitrary directions. Our experiments showed an increase of more than $25\%$ in performance, validating our hypothesis that \textit{R\&D} requires sufficient environmental observability and consistent demonstration data.

Failure modes for other tasks typically involved incorrect gripper action predictions. The main reason for this is that our rendered action representation does not include gripper actions, preventing the model from precisely understanding the implications of its actions. Additionally, we observed that the distribution of randomly collected demonstrations significantly impacts the performance of \textit{R\&D}. Specifically, we found that \textit{R\&D} struggles to complete tasks in regions of the workspace that have been infrequently observed in the demonstrations.

For the remaining experiments, we evaluate only the \textit{R\&D-AI} variant of our method and refer to it as \textit{R\&D}.

\subsection{Spatial Generalisation Experiments}
\label{sec:spatial_eval}

In our next set of experiments we systematically investigate our approach's spatial generalisation capabilities by studying its performance inside the convex hull the demonstrations. 

\begin{figure}[]
  \centering
  \includegraphics[width=\columnwidth]{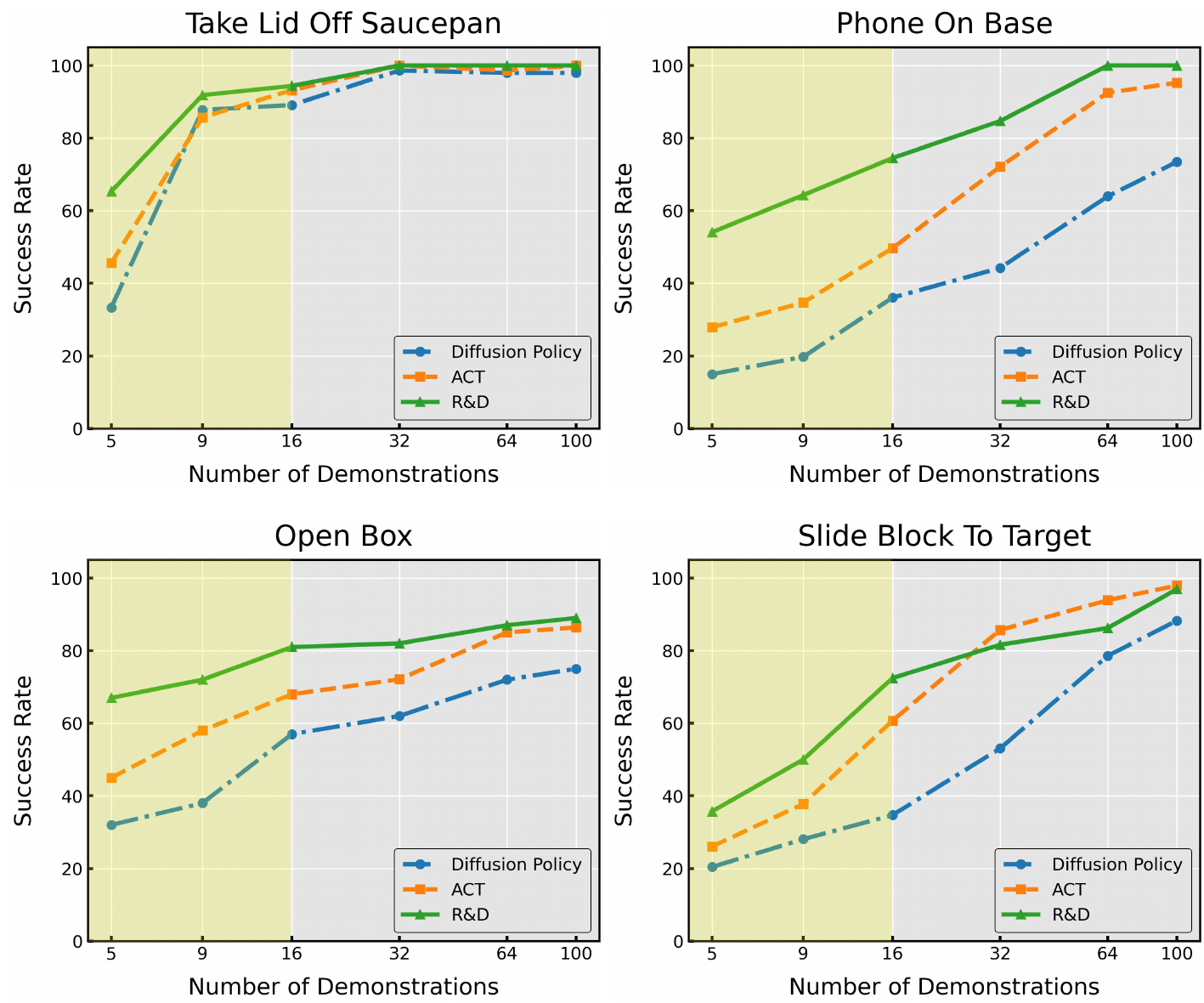} 
  \caption{Success rates \textit{vs} number of demonstrations for of \textit{R\&D} and the baselines. Increasing number of demonstrations covers the workspace in a uniform manner. Yellow-shaded regions indicate a low-data regime where the workspace is sparsely covered.}
  \label{fig:graphs}
\end{figure}

\textbf{Experimental Procedure.} For this set of experiments we selected $4$ tasks from RLBench that require significant spatial generalisation, namely: \textit{Open Box}, \textit{Lift Lid of Saucepan}, \textit{Place Phone on Base}, and \textit{Slide Block to Target}. 
We then define a demonstration collection procedure, which ensures that the poses of the objects can uniformly cover the workspace. 
Selected tasks do not involve big objects and thus can have bigger workspaces in which objects can be initialised, resulting in significantly different unseen object poses, requiring good spatial generalisation capabilities to complete these tasks from a limited number of demonstrations.
We start with a demonstration where the object is placed at the centre of the workspace and afterwards iteratively collect demonstrations where the pose of the object is selected such that it maximises the distance to all the other poses in the list of collected demonstrations. We do this using a greedy Furthest Point Sampling algorithm. In the beginning ($<6$ demos), only the edges of the workspace are covered and, at the limit, the whole workspace is covered densely. Due to workspace restrictions (for reachability purposes), no demonstrations in this set of experiments contain trajectories generated by RRT. 

We train all the models on different numbers of demonstrations collected in such a way and evaluate them on the whole workspace in a grid-like manner, varying both the position and orientation of the objects. Workspace dimensions, demonstration distributions and evaluation grid dimensions for each task can be found in the Appendix~\ref{sec:exp_details}.

\begin{figure}[]
  \centering
  \includegraphics[width=\columnwidth]{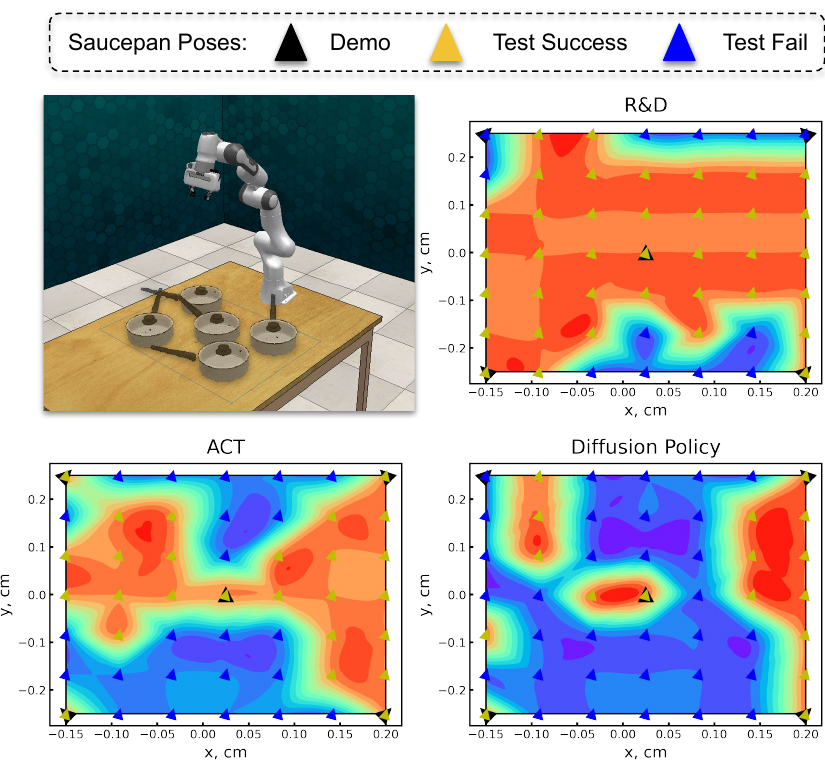}
  \caption{Outcomes of individual evaluation episodes of models trained on $5$ demonstrations (black triangles) for one of the Saucepan angles tested for the \textit{Lift Lid of Saucepan} task. The heat maps represent cubic interpolations between the sparse evaluations, with red and blue colours representing successes and failures, respectively.}
  \label{fig:exp_spatial}
\end{figure}

\textbf{Results \& Discussion.} Figure~\ref{fig:graphs} shows how the performance of \textit{R\&D} and the baselines change with increasing density of the workspace coverage, i.e. number of demos. We can see that, as expected all the methods significantly benefit from larger amounts of demonstrations. However, a stark contrast in performance is clearly visible between \textit{R\&D} and the baselines when only a few demonstrations are used to sparsely cover the workspace (left side of the graphs in Figure~\ref{fig:graphs}). Figure~\ref{fig:exp_spatial}, which shows the outcomes of individual evaluation episodes for one of the $\theta$ angles tested for the Saucepan task, clearly indicate that \textit{R\&D} has strong capabilities of interpolating within the convex hull of the demonstrations. This, again, shows the added value of the introduced inductive bias, achieved by aligning observation and action spaces. Please also note that the introduction of such an inductive bias did not diminish \textit{R\&D}'s capabilities of learning precise low-level actions needed to complete tasks such as \textit{Open Box}.
As the number of demonstrations increases, we observe a decreasing performance gap between \textit{R\&D} and the baselines (right side of the graphs in Figure~\ref{fig:graphs}). This trend is expected because, with enough demonstrations that densely cover the workspace, high-capacity models such as Diffusion Policy or ACT can implicitly learn the mapping between observation and action spaces with high accuracy. However, scaling curves still look favourable towards \textit{R\&D}.

\begin{figure}[]
  \centering
  \includegraphics[width=\columnwidth]{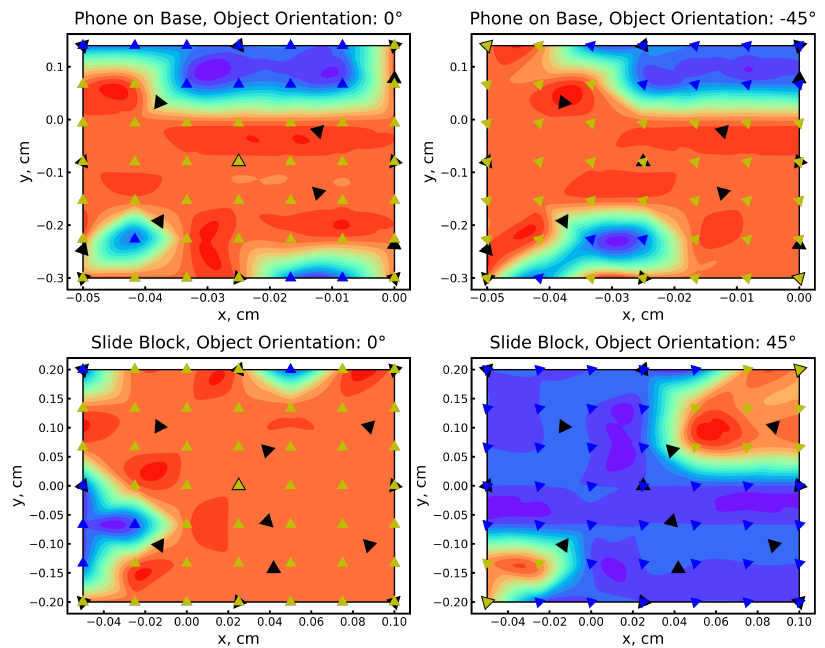}
  \caption{Outcomes of individual evaluation episodes of \textit{R\&D}, trained on $16$ demonstrations (black triangles) for \textit{Phone on Base} and \textit{Slide Block} tasks. The heat maps represent cubic interpolations between the sparse evaluations, with red and blue colours representing successes and failures, respectively.}
  \label{fig:exp_spatial_fail}
\end{figure}

\textbf{Failure Cases.} As in the previous set of experiments, a common failure mode of \textit{R\&D} was incorrect gripper action predictions; however, these failure modes become less frequent as the number of demonstrations increases. We also observed that with a sparsely covered workspace, the edges of the workspace (both in translation and rotation) were the most challenging. For instance, from Figure~\ref{fig:exp_spatial_fail}, we can see that \textit{R\&D} was mostly successful in pushing the block to the target when the object was initialised at an angle of $0$ (the middle value between orientation boundaries) but struggled significantly when the angle was $45$ degrees (the maximum value used when collecting demonstrations). Additionally, for the Phone on Base task, regardless of the orientation, \textit{R\&D} struggled in the furthest region of the workspace. Both of these cases require the robot to move significantly more compared to other regions of the workspace, allowing the errors in low-level action predictions to accumulate. However, as expected, performance in these challenging regions increases with additional demonstrations.

\subsection{Multi-Task Setting}
\label{sec:multi-task}
While our previous experiments involved training separate policies for each task, this set of experiments demonstrates \textit{R\&D}'s ability to learn multiple tasks using a single network. The purpose of this experiment is not to generalise to unseen tasks but rather to showcase that \textit{R\&D} is capable of absorbing data coming from different sources while remaining efficient.

\textbf{Experimental Procedure.} We adapt \textit{R\&D} (and the baselines) to include a goal specification as a learnable embedding and train a single network using data collected from $4$ RLBench tasks used in Section~\ref{sec:spatial_eval}. In this set of experiments, we use $20$ demonstrations that have been collected as described in Section~\ref{sec:sim_eval} and evaluate the trained models for $100$ episodes.

\begin{table}[]
\begin{tabular}{ll|clclclcl}
 &  & \multicolumn{2}{c}{Lift Lid} & \multicolumn{2}{c}{Open Box} & \multicolumn{2}{c}{Phone on Base} & Slide Block &  \\ \toprule
 & R\&D & \multicolumn{2}{c}{75 (-17)}  & \multicolumn{2}{c}{58 (+11)} & \multicolumn{2}{c}{40 (-23)} & 45 (-10) &  \\
 & DP   & \multicolumn{2}{c}{60 (-21)} & \multicolumn{2}{c}{41 (+2)} & \multicolumn{2}{c}{38 (-3)}  & 26 (-14) &  \\
 & ACT  & \multicolumn{2}{c}{67 (-22)} & \multicolumn{2}{c}{48 (+2)} & \multicolumn{2}{c}{42 (-8)}  & 40 (-18) & 
\end{tabular}
\caption{\label{tab:multi} Performance of \textit{R\&D-AI} and the baselines when a single model is trained on demonstrations from $4$ different tasks ($20$ demos each). We present success rates (\%) and differences in performance compared to a single-task setting.}
\end{table}

\textbf{Results \& Discussion.} Success rates (and the differences between the performance in a single-task setting) can be seen in Table~\ref{tab:multi}. We can see that, for all the methods, there is a decrease in performance compared to policies trained only on a single task. This is expected in a low-data regime due to several reasons. First, with limited amounts of diverse data, the learning problem becomes more complex. Second, the normalised action space expands and is no longer tailored for each task individually. Regardless, the performance of \textit{R\&D} remains favourable to the baselines and indicates that it is capable of learning how to complete multiple tasks using a single network.
Interestingly, all methods show an increased success rate for the open box task. We hypothesise that this is due to additional regularisation from more diverse data, which enhances the policy's robustness to small disturbances, such as minor collisions with the box or incomplete gripper closure.

\subsection{Ablations}
\label{sec:ablations}

In our last set of simulated experiments, we aim to justify our design choices and better understand the impact of various hyper-parameters on the performance of \textit{R\&D}. 

\begin{wraptable}{r}{0.4\columnwidth}
\begin{tabular}{l|r|}
\cline{2-2}
                                 & \multicolumn{1}{l|}{$\Delta$\%} \\ \hline
\multicolumn{1}{|l|}{Base}       & 0                          \\ \hline
\multicolumn{1}{|l|}{No-Texture} & -6                         \\ \hline
\multicolumn{1}{|l|}{No-Hist}    & -14                        \\ \hline
\multicolumn{1}{|l|}{Depth-6}    & -1                         \\ \hline
\multicolumn{1}{|l|}{Depth-12}   & -1                         \\ \hline
\multicolumn{1}{|l|}{Heads-8}    & -10                        \\ \hline
\multicolumn{1}{|l|}{Heads-20}   & 2                          \\ \hline
\end{tabular}

\caption{\label{tab:ablations} Performance change of the ablated variants compared to the base model.}
\end{wraptable}

\textbf{Experimental Procedure.} We train the variants of \textit{R\&D} with different sets of hyper-parameters and evaluate it on $4$ RLBench tasks used in Section~\ref{sec:spatial_eval}. We use $20$ demonstrations that have been collected as described in Section~\ref{sec:sim_eval} and evaluate the trained models for $100$ episodes. We compare these variants against the "base" model that is described in Section~\ref{sec:method} with hyper-parameters outlined in Section~\ref{sec:params}. We present the relative performance difference as percentage points.

\textbf{Different Variants.} \textit{R\&D-No-Texture} doesn't use textured colour when rendering the gripper; \textit{R\&D-No-Hist} uses only the current observation; \textit{R\&D-Depth-6} and \textit{R\&D-Depth-12} use $6$ and $12$ self attention layers respectively; \textit{R\&D-Heads-8} and \textit{R\&D-Heads-20} use $8$ and $20$ attention heads ($64$ dim per head) respectively;  \textit{R\&D-Steps-2}, \textit{R\&D-Steps-8}, and \textit{R\&D-Steps-16} use different number of diffusion time steps during inference;

\textbf{Results \& Discussion.} From Table~\ref{tab:ablations}, we can see that including more than just the current observation makes a huge different in performance. Subsequently, adding textures to renders to disambiguate the full 6D pose of the gripper is also important. While the depth of the transformer doesn't seem to greatly impact the networks capabilities, its width has a significant effect on the model's performance. It is important to note that increasing the model's capacity comes with an increase in computational requirements. Finally, the number of diffusion time steps doesn't seem to change the performance of our approach, given they are above a certain threshold. Additional discussion on hyper-parameters and what we found to work and not work, can be found in the Appendix~\ref{sec:trials}.

\subsection{Real-World Deployment}
\label{sec:real_world}

In our last set of experiments, we investigate \textit{R\&D}'s applicability in solving everyday tasks in the real world.

\begin{figure}[]
  \centering
  \includegraphics[width=\columnwidth]{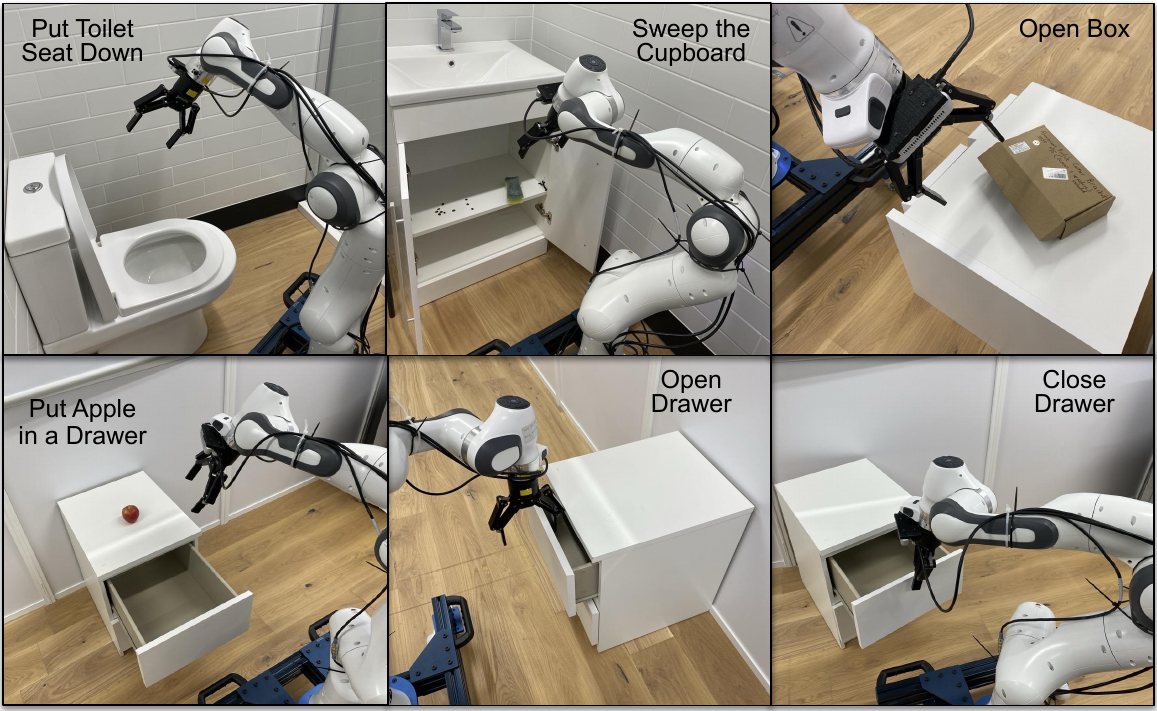} 
  \caption{Six everyday tasks used for our real-world evaluation.}
  \label{fig:rw_tasks}
\end{figure}

\textbf{Experimental Procedure.} We evaluate \textit{R\&D} using a real Franka Emika Panda robot equipped with a Robotiq 2F-140 gripper on $6$ everyday tasks seen in Figure~\ref{fig:rw_tasks}. For each task, we collect $20$ demonstrations using $2$ robots controlled via joint mirroring. During demonstration collection, relative poses between the robot and the objects in the environments are randomised. 
The relative poses between the robot and the objects are randomised by either moving the objects with respect to the robot (e.g. for the open box task, we move the drawer and the box on top of it) or moving the robot with respect to a static object using a wheeled base (e.g. for the toilet seat down task). The positioning of the objects (or the robot) is done such that the external camera can see the object of interest and the task is kinematically feasible, limiting the amount of randomisation available.
We use $2$ calibrated RealSense D415 cameras to capture RGB-only observations of the environment: one external (left shoulder) camera and one mounted on the wrist of the robot. Following a thorough and systematic evaluation of \textit{R\&D} in simulation, we focus on qualitatively showing its applicability in the real world and run $10$ evaluations for each task, randomising the poses of the objects in the environment every time. 
During testing, the robot's relative position to objects is selected within the convex hull of poses observed during demonstrations, assessing the model's interpolation rather than extrapolation capabilities.
In our experiments, we used a standard end-effector position controller to execute the actions predicted by \textit{R\&D}.

\begin{wraptable}{r}{0.5\columnwidth}
\begin{tabular}{|l|l|}
\hline
Toilet Seat Down & 10 / 10 \\ \hline
Sweep Cupboard & 8 / 10 \\ \hline
Place Apple & 8 / 10 \\ \hline
Open Box & 10 / 10 \\ \hline
Open Drawer & 10 / 10 \\ \hline
Close Drawer & 10 / 10 \\ \hline
\end{tabular}%
\caption{\label{tab:rw} Performance of \textit{R\&D} on six everyday tasks.}
\end{wraptable}

\textbf{Results \& Discussion.} From Table~\ref{tab:rw}, we can see that \textit{R\&D} is capable of completing everyday tasks in the real world, where different noise sources such as imperfect camera calibration are present. However, failure cases were observed for the \textit{Place Apple in a Drawer} and \textit{Sweep Cupboard} tasks. For the former, this happened when the robot moved its gripper in a position obstructing the apple without it being in view for the wrist camera. We hypothesise that this issue could be addressed by adding more past observations to the model or using multiple external cameras. For the latter, the compliant grasping of the sponge sometimes resulted in not all of the coffee beans being swept. This is expected from a policy trained without corrective actions and access to force measurements. Interestingly, we also observed that \textit{R\&D} was still able to complete these tasks with a high success rate ($\sim 70 \%$) with the presence of bright distractors or even multiple objects of the same category in the scene (e.g. multiple apples) without any data augmentation or additional techniques.

\section{Limitations \& Future Work} 
\label{sec:limitations}

\textbf{Limitations.} Although our experiments show promising results, especially when it comes to spatial generalisation, task understanding and generalisation to distractors, \textit{R\&D} still has its limitations. First of all, inference involves an iterative process that includes multiple rounds of rendering and forward propagation through the whole model, which can be computationally expensive. Secondly, it relies on camera calibration which can be unavailable in some settings. Moreover, as evident from its performance on \textit{Open Drawer} task in our simulation experiments, it is not well versed when dealing with tasks with severe occlusions and inconsistent data. Furthermore, high-dimensional input and output spaces of the network can make the training difficult, necessitating evaluating multiple checkpoints to achieve the best performance. Finally, due to the high number of possible method variations and hyper-parameters, it can be difficult to find the optimal set of parameters for a given task.  

\textbf{Future Work.} Despite the aforementioned limitations of our proposed method, we are excited about the possible lines for work that would address them and its general potential as a universal way to jointly represent RGB observations and actions. Particularly, we will be looking into ways to extend such a representation to include the full configuration of the robot (including gripper actions), improving its computational requirements by leveraging different network architectures, and alternative ways of fusing predictions expressed in different spaces. Additionally, as we express robot actions in the image space, integrating such a representation with image foundation models, such as DINO~\cite{caron2021emerging}, holds a huge potential.

\section{Conclusion} 
\label{sec:conclusion}

In this work, we proposed \textit{Render and Diffuse} a novel method that represents RGB observations and low-level robot action inside a unified image space. Using such a representation together with a learned denoising process were able to achieve promising results in spatial generalisation as well as generalisation to distractor objects. We thoroughly evaluated our proposed approach in simulated environments and showcased its applicability in completing everyday tasks in the real-world. Overall, we are excited about this direction and the potential it holds.



\bibliographystyle{plainnat}
\bibliography{references}

\clearpage
\appendix

\section*{Things that Increased / Decreased Performance}
\label{sec:trials}

\begin{itemize}
    \item \textbf{Normalisation.} We found that normalising the outputs to be $[-1, 1]$ is a crucial step, without which the model is not capable of learning anything meaningful.
    \item \textbf{Decreasing the Learning Rate.} As we are using small batch sizes for computational reasons, a high learning rate (e.g. $1e^{-3}$) introduces significant training instabilities. Setting the learning rate too small, however, increases the time required for the network to converge. We settled on a learning rate of $1e^{-4}$.
    \item \textbf{Decreasing patch size.} We tried using a larger patch size of $32 \times 32$ to increase the computational efficiency of our method. We found it to drastically reduce the performance. Decreasing the patch size to $8 \times 8$ showed improved performance for some of the tasks, however, it came with a significant increase in computational requirements.
    \item \textbf{MSE Loss.} Changing the loss function from $L1$ to $MSE$ resulted a decrease in performance.
    \item \textbf{L1 Loss for the Gripper Actions.} We tried including the gripper actions in the diffusion process and using $L1$ loss (same as for the $6D$ actions similar to \cite{chi2023diffusion_policy}). The performance of the model decreased significantly, with most of the failure cases being wrong gripper action predictions.
    \item \textbf{Predicting Gripper Actions in Image Space.} We tried predicting the gripper actions in the image space as an additional output channel and averaging it to produce the final gripper action predictions. We found that this prediction strategy did not work and hindered the model's ability to learn both precise $6D$ actions and binary gripper actions.
    \item \textbf{Using ResNet Encoders.} To reduce the dimensionality of the inputs, we tried processing images with ResNet encoders~\cite{he2016deep} before inputting them into a Transformer. We observed a significant drop in performance.
    \item \textbf{Using UNet Model.} Initially, we tried using a standard UNet~\cite{ronneberger2015u} architecture, fusing information from multiple cameras and time steps in the latent space. The performance of the model was significantly worse, indicating that the transformer model is better equipped for fusing information from different sources.
    \item \textbf{Deterministic Sampling.} We tried starting the diffusion process deterministically, ensuring that initially, all renders of the gripper were visible by the wrist camera. We found that, on average, this approach reduced performance.
    \item \textbf{Adding Proprioceptive Inputs.} Instead of providing just the gripper state information, we tried adding the current $6D$ pose of the gripper to the input of the model. We observed no major change in performance.
    \item \textbf{Using Exponential Moving Average (EMA).} We observed no major change in performance when using EMA to increase the stability of training.
    \item \textbf{View-Masking.} We experimented with randomly masking out information from different camera views during training. Performance increased for some tasks and decreased for others. We believe that view/information masking holds significant potential for increasing performance, and more investigation is needed.
\end{itemize}

\section*{Experimental Details}
\label{sec:exp_details}

\subsection*{Evaluation in Simulated Environments}
\label{sec:sim_eval_details}

\begin{figure}[ht!]
  \centering
  \includegraphics[width=\columnwidth]{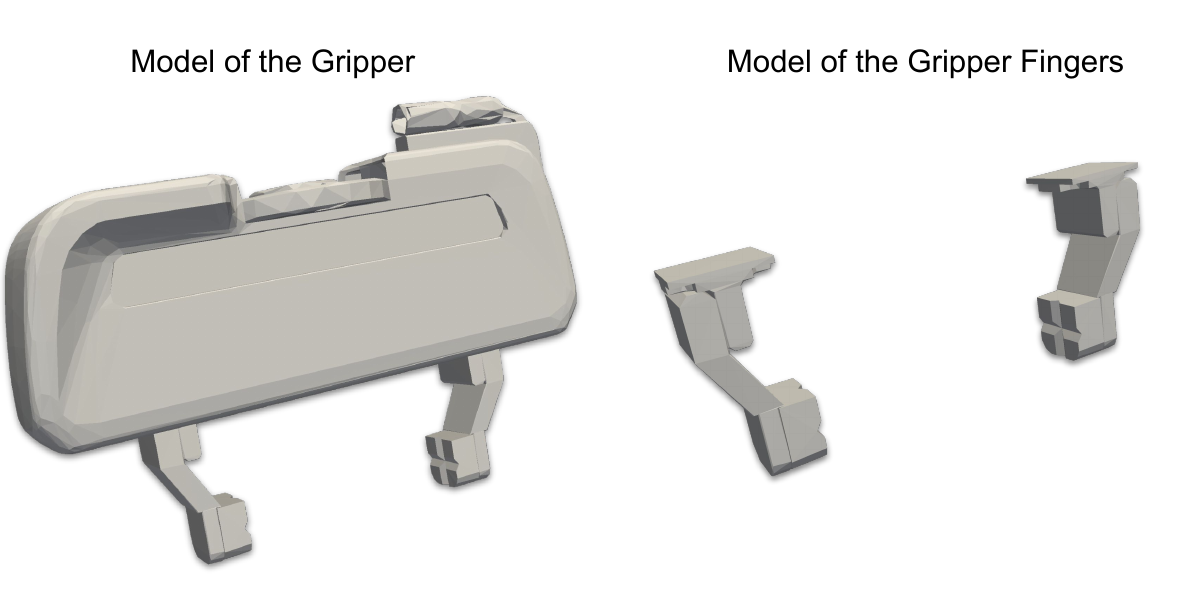} 
  \caption{Visualisation of the 3D meshes used in our simulated experiments.}
  \label{fig:franka_gripper}
\end{figure}

\textbf{Gripper's 3D Model.} For our simulated experiments, we used a Franka Emika Panda robot and rendered its open gripper to create our rendered action representation. For the wrist camera, we use only its fingers to avoid severe occlusions. Visualisations of the 3D models we used for rendering can be seen in Figure~\ref{fig:franka_gripper}.

\textbf{Alterations of RLBench Tasks.} To conduct our experiments thoroughly and systematically, we had to alter some of the RLBench tasks. First, we changed the starting configuration of the robot so that the whole gripper is in view from the front camera. We did this to study the performance of one of the \textit{R\&D} variants that relies solely on predictions made in the image space - \textit{R\&D-I}. Second, for some tasks, we restricted the orientation of the objects in the environment. This was necessary because, otherwise, the majority of the collected demonstrations included trajectories generated with the RRT-Connect planner, which removed the correlation between the observations and low-level actions, making it extremely hard to learn well-performing policies from a small number of demonstrations. These tasks included: \textit{Place Phone on Base}, \textit{Slide Block to Target}, and \textit{Close Laptop} ($-45^{\circ} $ - $45^{\circ}$).

\subsection*{Spatial Generalisation Experiments}
\label{sec:spatial_generalisation_details}

\begin{figure}[ht!]
  \centering
  \includegraphics[width=\columnwidth]{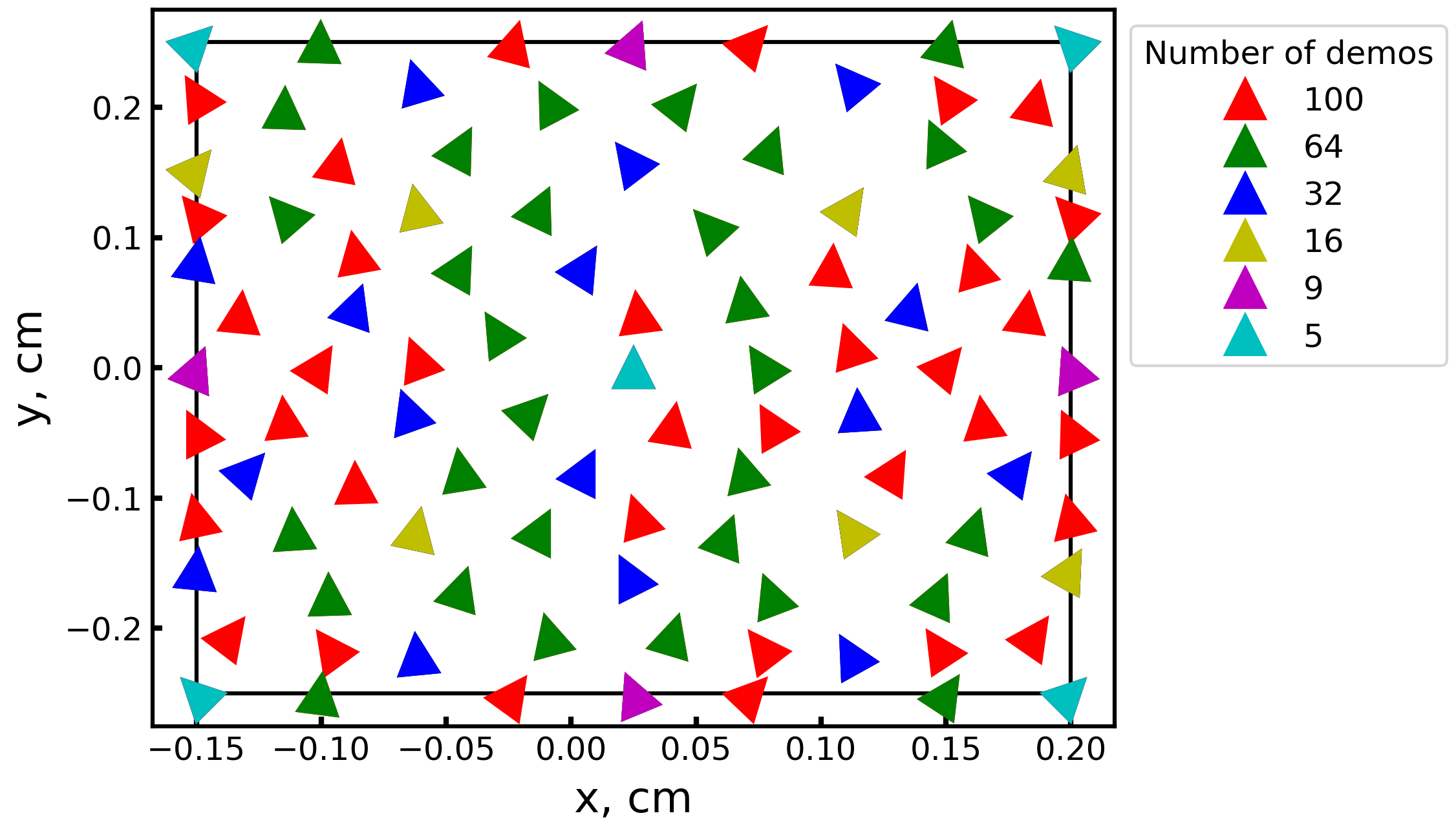} 
  \caption{Visualisation of the workspace coverage with an increasing number of demonstrations for the \textit{Lift Lid of Saucepan} task. Triangle orientation represents the orientation of the saucepan. Different colours indicate at what stage these demonstrations were added.}
  \label{fig:demos}
\end{figure}

\textbf{Demo Sampling Procedure.}
To test our model's spatial generalisation capabilities and cover the workspace with increasing levels of density, we utilised a Furthest Point Sampling algorithm. We greedily added demonstrations to the dataset that are the most different from the ones already collected. This strategy results in initially sparse workspace coverage (cyan triangles in Figure~\ref{fig:demos}) that becomes denser as more demos are added.

\textbf{Workspace Dimensions.}
Here we describe the workspace dimensions used in our spatial generalisation experiments for each of the considered tasks. These dimensions were chosen to be as big as possible while still making sure that the tasks can be completed within the whole workspace (due to kinematic constraints). We present them as $x \times y \times \theta$, where $\theta$ is the angle of rotation around the $z$ axis.

\begin{itemize}
    \item \textit{Lift Lid of Saucepan:} $35 cm \times 44 cm \times 90^{\circ}$
    \item \textit{Place Phone on Base:} $5 cm \times 44 cm \times 90^{\circ}$
    \item \textit{Open Box:} $5 cm \times 20 cm \times 45^{\circ}$
    \item \textit{Slide Block to Target:} $15 cm \times 40 cm \times 90^{\circ}$
\end{itemize}

\subsection*{Real-World Deployment}
\label{sec:real_world_details}
\begin{figure}[ht!]
  \centering
  \includegraphics[width=\columnwidth]{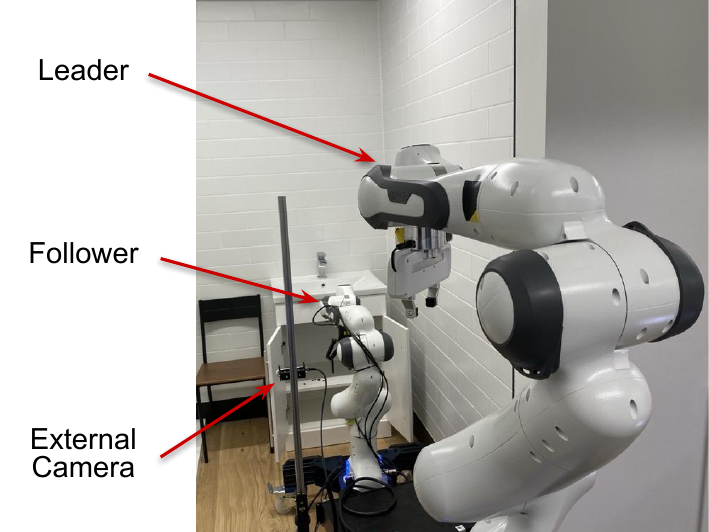} 
  \caption{Real world setup used to collect demonstrations via joint mirroring.}
  \label{fig:setup}
\end{figure}

In the real world, we collect demonstrations for different tasks using joint mirroring. A \textit{Leader} robot is moved by the operator, and the \textit{Follower} robot mirrors its movement in joint space. Robot states and observations from an external camera and a camera mounted on the wrist are recorded during this process, creating a dataset of demonstrations. The setup used can be seen in Figure~\ref{fig:setup}.

\begin{figure}[ht!]
  \centering
  \includegraphics[width=\columnwidth]{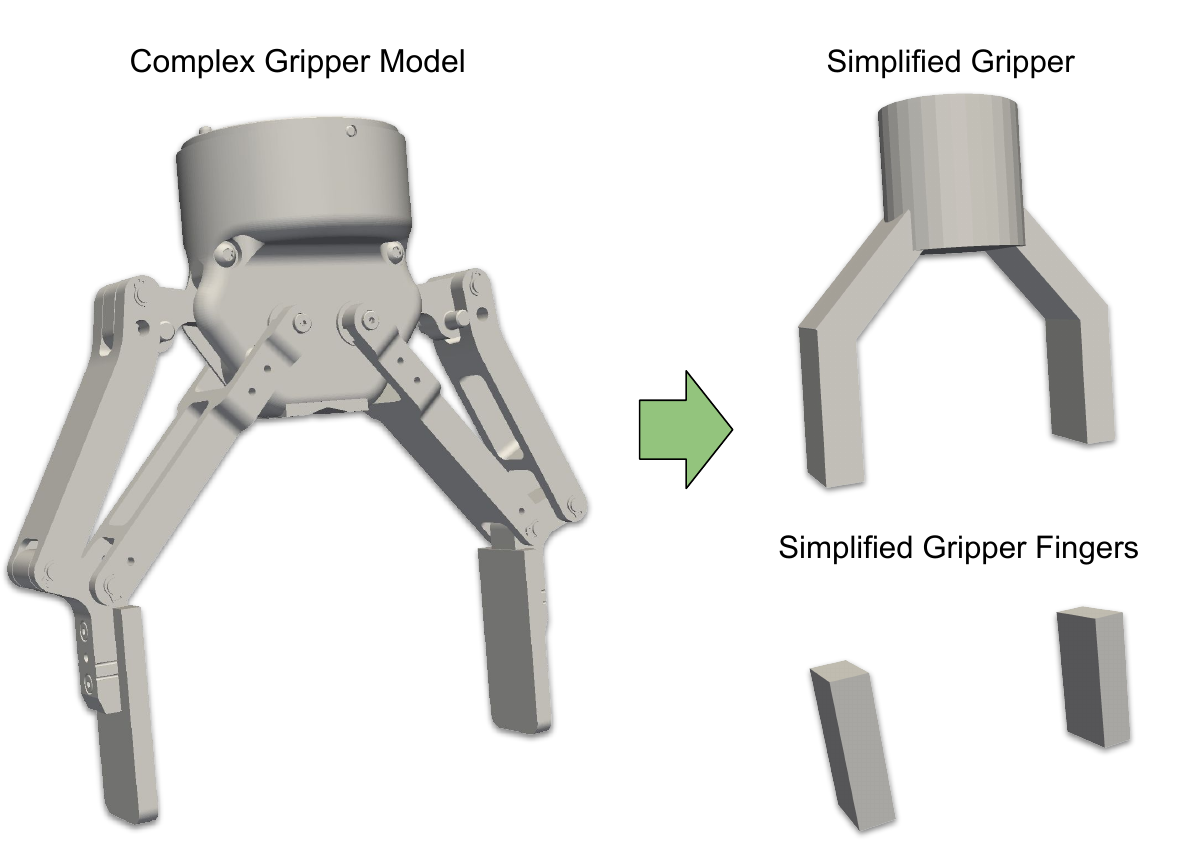} 
  \caption{Visualisation of the simplification process of the Robotiq 2F-140 gripper. Simplified meshes were used in our real-world experiments.}
  \label{fig:real_gripper}
\end{figure}

\textbf{Gripper's 3D Model.} For our real-world experiments, we employ a Franka Emika Panda robot equipped with a Robotiq 2F-140 gripper. Due to the complexity of the gripper's mesh, rendering times increase without providing significant additional information at the resolution we are using ($128 \times 128$). Therefore, we opted for a heavily simplified 3D model of the gripper, which is depicted in Figure~\ref{fig:real_gripper}.

\textbf{Task Definitions.}
\begin{itemize}
    \item \textit{Toilet Seat Down.} The goal is to put the toilet seat down, making sure it is fully down by first flipping it and then pushing it down. Success is considered if the toilet seat ends up fully closed.
    \item \textit{Sweep Cupboard.} The goal is to pick up the sponge and sweep the coffee beans in the cupboard to one side. Success is considered if coffee beans end up in a pile on the left side of the cupboard.
    \item \textit{Place Apple.} The goal is to pick up an apple and place it inside the open drawer. Success is considered if the apple ends up inside the drawer.
    \item \textit{Open Box.} The goal is to fully open a cardboard box. This includes flipping the lid open and pushing it down if it is not fully open. Success is considered if the box ends up completely open.
    \item \textit{Open Drawer.} The goal is to slide a partially open drawer such that it is completely open. Success is considered if the drawer ends up being open.
    \item \textit{Close Drawer.} The goal is to slide a fully open drawer such that it is completely closed. Success is considered if the drawer ends up being completely closed.
\end{itemize}

\end{document}